\documentclass[journal,comsoc]{ieeetran}
\usepackage{cite}
\usepackage{amsmath,amssymb,amsfonts}
\usepackage{algorithmic}
\usepackage{graphicx}
\usepackage{multirow}
\usepackage{subfigure} 
\usepackage{textcomp}
\usepackage{url}
\newcommand\norm[1]{\left\lVert#1\right\rVert}
\def\BibTeX{{\rm B\kern-.05em{\sc i\kern-.025em b}\kern-.08em
    T\kern-.1667em\lower.7ex\hbox{E}\kern-.125emX}}
\begin{document}
\title{Road Damage Detection Based on Unsupervised Disparity Map Segmentation}
\author{Rui Fan, \IEEEmembership{Member, IEEE}, and Ming Liu, \IEEEmembership{Senior Member, IEEE}
	\vspace{-3em}	
\thanks{\textit{Corresponding author: Ming Liu}}
\thanks{The authors are with the Robotics and Multi-Perception Laboratory, Robotics Institute, the Hong Kong University of Science and Technology, Hong Kong (e-mail: rui.fan@ieee.org; eelium@ust.hk).}
\thanks{This work was supported by the National Natural Science Foundation of China, under grant No. U1713211, the Research Grant Council of Hong Kong SAR Government, China, under Project No. 11210017, No. 21202816, and the Shenzhen Science, Technology and Innovation Commission (SZSTI) under grant JCYJ20160428154842603, awarded to Prof. Ming Liu.}
}

\maketitle

\begin{abstract}
This paper presents a novel road damage detection algorithm based on unsupervised disparity map segmentation. Firstly, a disparity map is transformed by minimizing an energy function with respect to stereo rig roll angle and road disparity projection model. Instead of solving this energy minimization problem using non-linear optimization techniques, we directly find its numerical solution. The transformed disparity map is then segmented using Otus's thresholding method, and the damaged road areas can be extracted. The proposed algorithm requires no parameters when detecting road damage. The experimental results illustrate that our proposed algorithm performs both accurately and efficiently. The pixel-level road damage detection accuracy is approximately $\boldsymbol{97.56\%}$. 
\end{abstract}

\begin{IEEEkeywords}
road damage detection, disparity map segmentation, stereo rig roll angle, road disparity projection model, numerical solution.
\end{IEEEkeywords}

\vspace{-1em}
\section*{Supplementary Materials}
The source code is publicly available at: \url{https://github.com/ruirangerfan/unsupervised_disparity_map_segmentation.git}

\section{Introduction}
\label{sec.introduction}
Road damage, notably pothole or crack, is not just an inconvenience, but also a safety hazard \cite{fan_thesis}. Road damage is regularly detected by certified inspectors \cite{Kim2014}. This process is, however, cumbersome, costly and time-consuming \cite{Fan2018}. Furthermore, the road damage detection results are always subjective, as they depend entirely on the inspectors' experience \cite{Koch2015}. Therefore, there is an ever-increasing need for automated road condition assessment systems that can recognize and localize road damage both efficiently and objectively \cite{Mathavan2015}. The rest of this section presents the state of the art in road damage detection and highlights the motivation, contributions and structure of this paper.
\subsection{State-of-the-Art Road Damage Detection Methods}
\label{sec.SoA}
Over the past decade, passive and active sensing technologies have been extensively used to acquire 2D/3D road data \cite{Koch2015}. 2D color/gray-scale road images are typically captured by digital cameras \cite{Koch2011}, while 3D road data, e.g., road point cloud or road depth/disparity map, are supplied by laser scanners \cite{Tsai2017}, Microsoft Kinect sensors \cite{Jahanshahi2012}, or passive sensors (i.e., a single movable camera \cite{jog2012pothole} or an array of synchronized cameras \cite{fan2018roadreconstruction}). The state-of-the-art road damage detection methods can be classified as either 3D road surface modeling-based \cite{fan2019pothole} or 2D image analysis-based. 
The former commonly fits a quadratic surface to the raw 3D road data and detect the damaged road areas by comparing the difference between the raw data and the modeled road surface \cite{fan2019pothole}. 

On the other hand, 2D image analysis-based road damage detection methods can be grouped into two categories: computer vision-based \cite{Li2016, fan2018novel, buza2017unsupervised, Koch2011, Tsai2017, fan2019real} and machine learning-based \cite{Cord2012, cha2017deep, bhatia2019convolutional,wu2019road}. The former typically pre-processes a 2D image, i.e., an RGB/gray-scale image or a depth/disparity map, using some image processing techniques, e.g., various image filters, to reduce image noise and enhance road damage outline \cite{Li2016, fan2018novel}. The pre-processed image is then segmented using some thresholding methods, such as Otsu \cite{buza2017unsupervised}, triangle \cite{Koch2011} or watershed \cite{Tsai2017}, to extract damaged road areas. In \cite{fan2018novel}, we proposed a disparity transformation algorithm which can better distinguish between damaged and undamaged road areas. The transformation parameters were estimated by minimizing an energy function using golden section search (GSS) and dynamic programming (DP). Recently, we proposed to minimize the aforementioned energy function using gradient descent (GD), which has shown a better efficiency \cite{fan2019real}.

With recent advances in supervised learning, deep convolutional neural networks (CNNs) have been used for road image classification and semantic road image segmentation. For example, Cha et al. \cite{cha2017deep} cropped the RGB images into a group of squared image patches and labeled them as either positive  or negative ones. The labeled training data were then used to train a CNN for road image patch classification \cite{cha2017deep}. In \cite{bhatia2019convolutional}, the authors utilized thermal images to train a residual network (ResNet) \cite{he2016deep} for road image classification. Furthermore, Wu et al. \cite{wu2019road} developed a robust road image segmentation system based on DeepLabv3+ \cite{chen2017rethinking}, which employs atrous convolution along with upsampled filters to extract dense feature maps and to capture long-range context.

\vspace{-1em}
\subsection{Motivation}
\label{sec.motivation}
Currently, laser scanning is still the main technology used for 3D road data acquisition, while other technologies, such as passive sensing, are under-utilized \cite{Mathavan2015}. However, the  long-term maintenance of such laser scanners is still very expensive \cite{Koch2011}. Furthermore, Microsoft Kinect sensors were initially designed for indoor use, and they suffer greatly from infra-red saturation in direct sunlight \cite{Abbas2012}. Therefore, the trend of 3D road data acquisition is to utilize digital cameras, notably stereo cameras.  

For 3D road surface modeling-based methods, finding the best parameters is very challenging, as the parameters they select cannot be applied to all cases \cite{fan2019pothole}. On the other hand, computer vision-based methods can recognize road damage with low computational complexity, but the achieved detection accuracy is still far from satisfactory \cite{Koch2015}. For machine learning-based methods, training a road image classification/segmentation neural network using supervised learning requires a large amount of labeled training data, and producing such data can be a long and labor-intensive task \cite{Koch2015}. Moreover, color/gray-scale image segmentation is always severally affected by various environment factors, notably illumination conditions, but disparity/depth map segmentation is not subject to such environment factors \cite{Salari2011}. Therefore, there is a strong motivation to explore unsupervised disparity map segmentation method for road damage detection.

\subsection{Novel Contributions}
\label{sec.contributions}

In this paper, we present a real-time road damage detection algorithm based on unsupervised disparity map segmentation. The proposed algorithm is developed from our previous work \cite{ fan2019pothole}, where road disparity maps were transformed to better distinguish between damaged and undamaged road areas. Instead of estimating the transformation parameters using non-linear optimization methods, such as GSS-DP and GD, we directly find the numerical solution for the energy minimization problem stated in \cite{fan2019pothole}. The proposed algorithm is capable of segmenting dense disparity maps for road damage detection without setting any parameters. Furthermore, the stereo rig roll angle can be accurately estimated from disparity maps, which enables our method to be utilized for vehicle state estimation. We also believe this algorithm can be utilized to automatically label training data for road damage detection.
\begin{figure}[!t]
	\centering
	\includegraphics[width=0.40\textwidth]{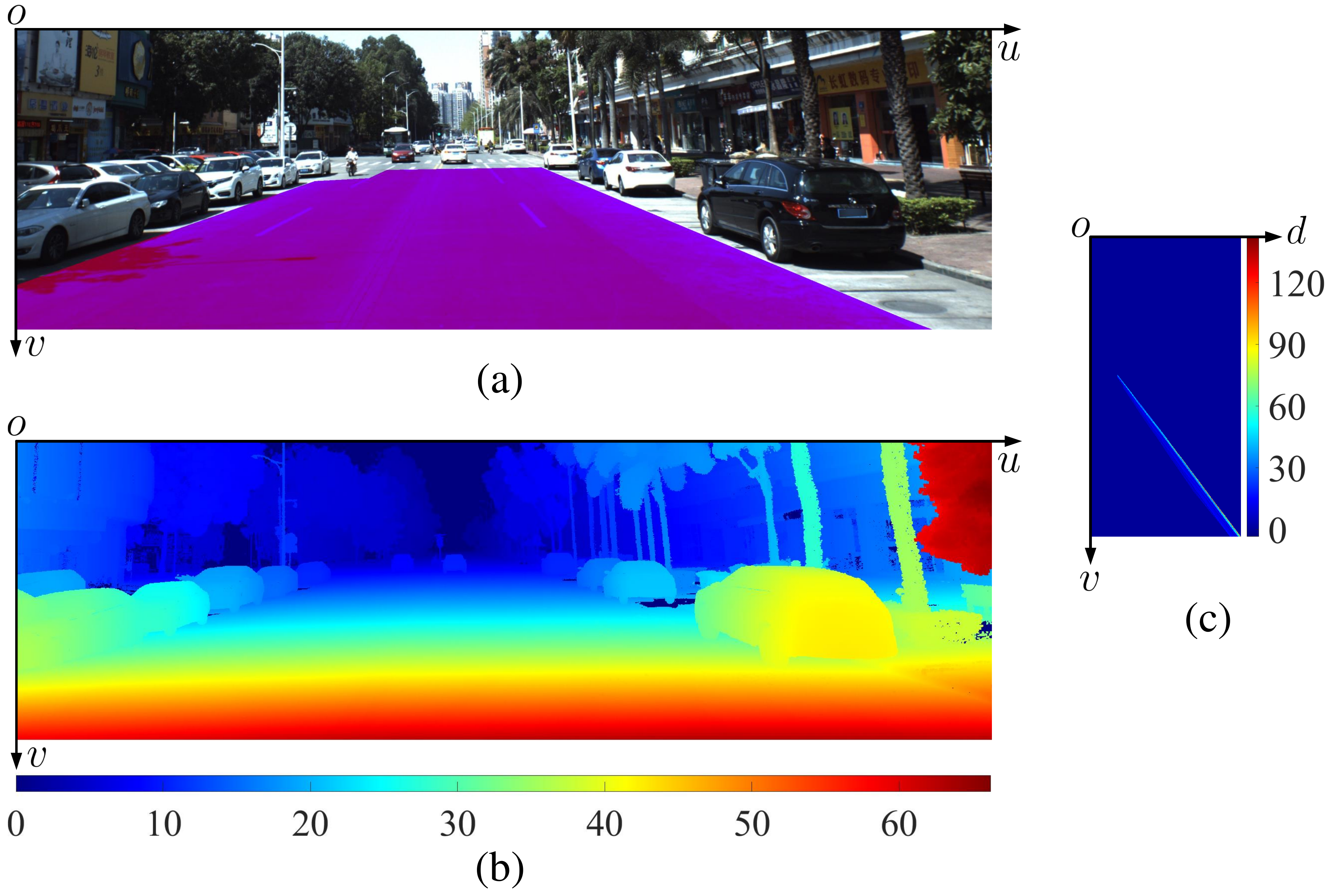}
	\caption{(a) left stereo image, where the area in purple is our manually labeled road region; (b) disparity map; (c) v-disparity image. }
	\label{fig.v_disp}
\end{figure}

\begin{figure}[!t]
	\begin{center}
		\centering
		\subfigure[]
		{
			\includegraphics[width=0.18\textwidth]{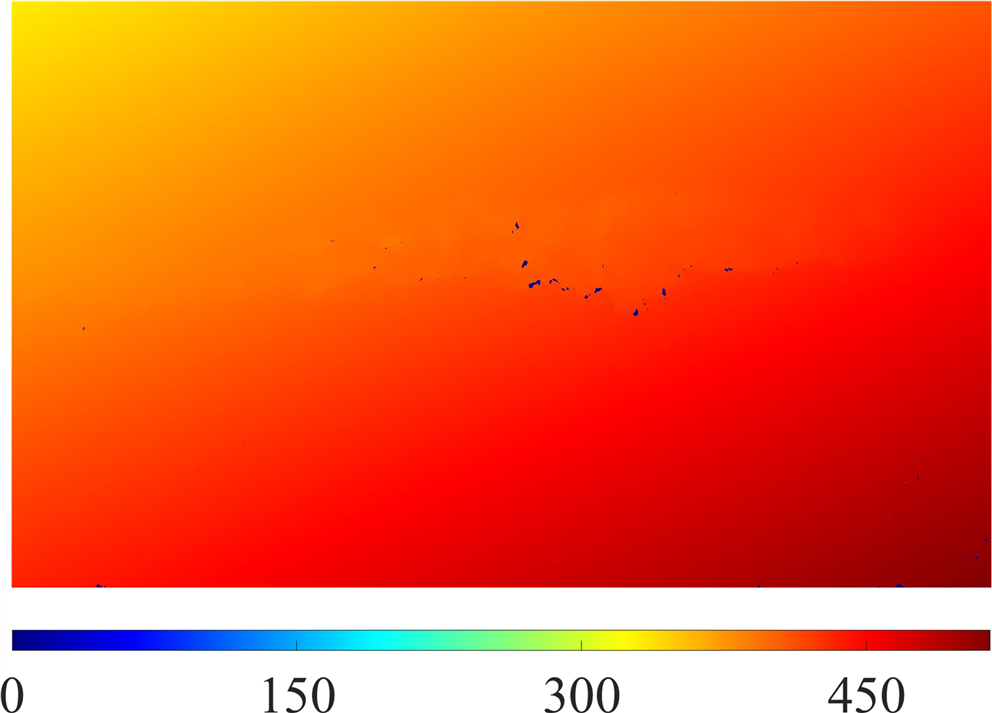}
			\label{fig.disp}
		}
		\subfigure[]
		{
			\includegraphics[width=0.18\textwidth]{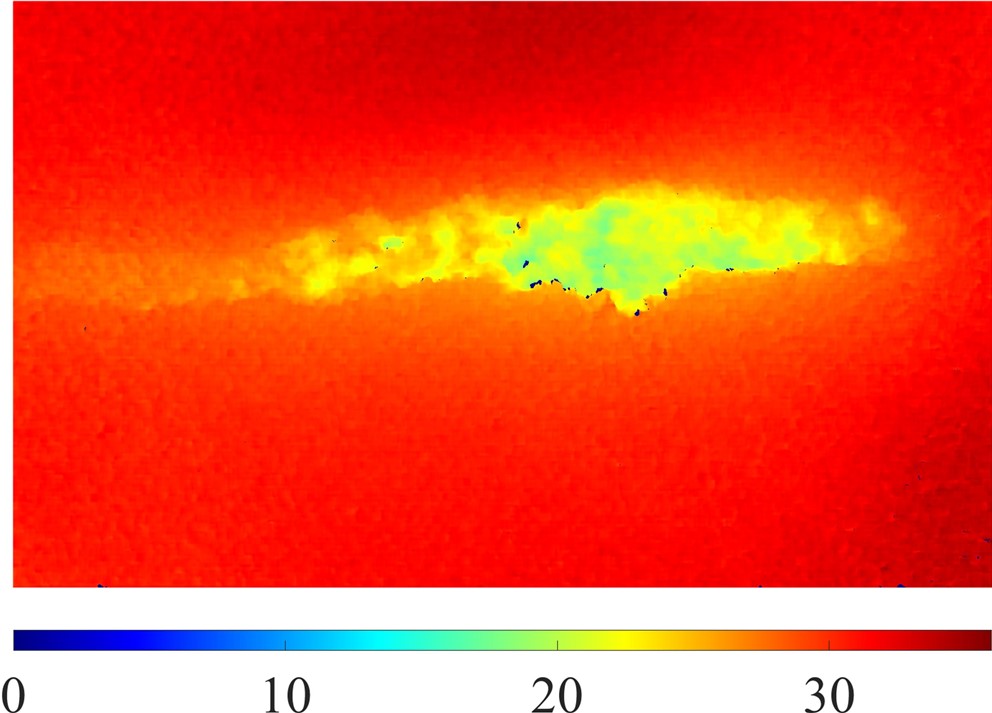}
			\label{fig.trans_disp}
		}\\
		\subfigure[]
		{
			\includegraphics[width=0.18\textwidth]{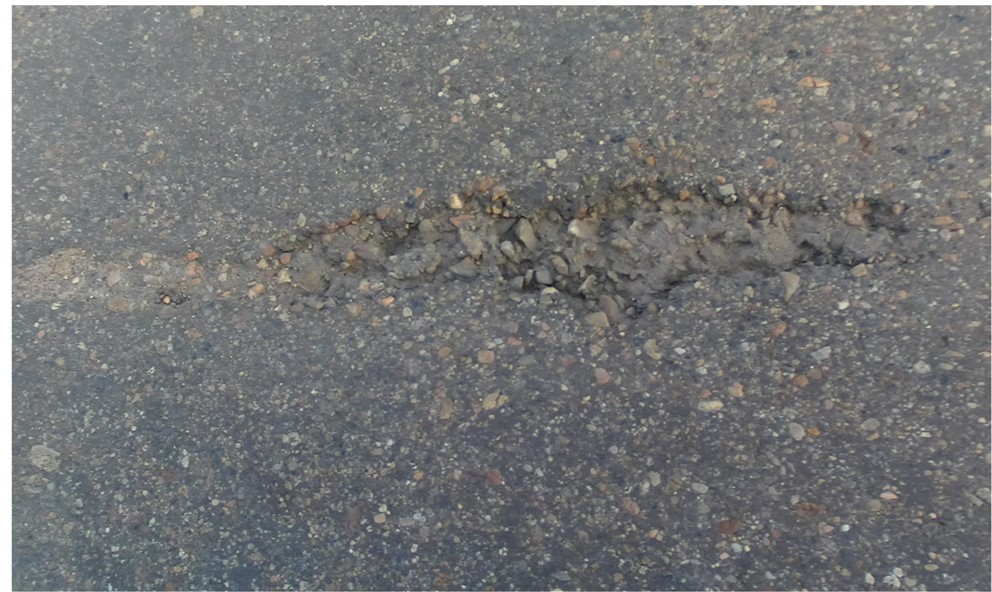}
			\label{fig.rgb}
		}
		\subfigure[]
		{
			\includegraphics[width=0.18\textwidth]{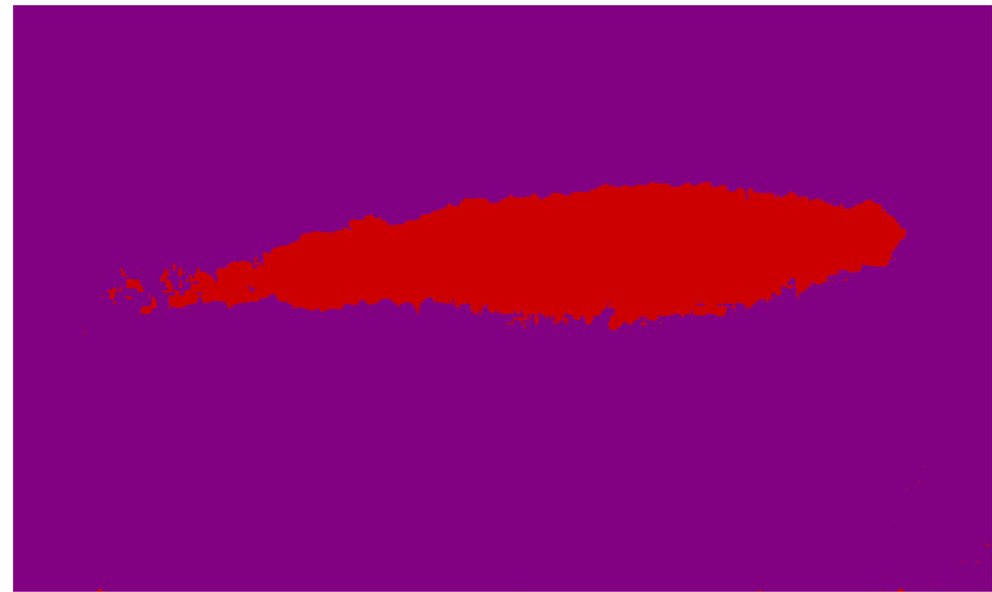}
			\label{fig.seg}
		}
		\caption{Unsupervised disparity map segmentation:  (a) original disparity map; (b) transformed disparity map; (c); left stereo image; (d) road damage detection result, where the regions in purple and red are undamaged and damaged road areas, respectively.}
	\end{center}
		\vspace{-2em}
\end{figure}
\subsection{Paper Structure}
\label{sec.paper_outline}

The remainder of this paper is organized as follows: Section \ref{sec.preliminaries} introduces v-disparity image and road disparity projection model. Section \ref{sec.algorithm_description} presents the proposed unsupervised road damage detection algorithm. The experimental results are illustrated and the algorithm performance is discussed in Section \ref{sec.experimental_results}. Finally, Section \ref{sec.conclusion} summaries the paper. 

\section{Preliminaries}
\label{sec.preliminaries}

Since Labayrade and Aubert \cite{Labayrade2003} introduced the concept of ``v-disparity image'' in 2003, disparity map has been widely used for road region extraction \cite{fan2018lane}. An example left stereo image and its corresponding dense disparity map are shown in Fig. \ref{fig.v_disp}(a) and \ref{fig.v_disp}(b), respectively, where the purple area in Fig. \ref{fig.v_disp}(a) is our manually labeled road region. By computing the disparity histogram with respect to each image row, a v-disparity image can be created \cite{Ozgunalp2017}, as shown in Fig. \ref{fig.v_disp}(c).

Since the road surface is generally considered as a ground plane, for a stereo rig whose baseline is perfectly parallel to the road surface, its roll angle $\theta$ equals $0$, and the disparities on each row have similar values \cite{fan2018novel}, as shown in Fig. \ref{fig.v_disp}(b). Therefore, the projections of road disparities on the v-disparity image can be represented by a linear model \cite{Hu2005}:
\begin{equation}
f(\mathbf{p})=a_0+a_1 v,
\label{eq.linear_model}
\end{equation}
where $\mathbf{a}=[a_0,a_1]^\top$ stores the coefficients of the linear model, and $\mathbf{p}=[u,v]^\top$ is a pixel in the disparity map. $\mathbf{a}$ can be estimated by minimizing the following energy:
\begin{equation}
E=\norm{\mathbf{d}-\mathbf{V}\mathbf{a}}^2_2,
\label{eq.E}
\end{equation}
where $\mathbf{d}=[d_1,d_2,\cdots,d_n]^\top$ stores the disparity values. $\mathbf{V}=[\mathbf{1}_{n}, \ \mathbf{v}]$, where $\mathbf{1}_k$ represents a $k\times1$ vector of ones and $\mathbf{v}=[v_1,v_2,\cdots,v_n]^\top$. The above energy minimization problem has a closed form solution:
\begin{equation}
\mathbf{a}=(\mathbf{V}^\top\mathbf{V})^{-1}\mathbf{V}^\top\mathbf{d}.
\label{eq.alpha}
\end{equation}
Plugging (\ref{eq.alpha}) into (\ref{eq.E}) obtains the minimum energy:
\begin{equation}
E_{\text{min}}=\mathbf{d}^\top\mathbf{d}-\mathbf{d}^\top\mathbf{V}(\mathbf{V}^\top\mathbf{V})^{-1}\mathbf{V}^\top\mathbf{d}.
\label{eq.E_min}
\end{equation}

\section{Algorithm Description}
\label{sec.algorithm_description}
However, when the stereo rig baseline is not parallel to the road surface, a non-zero roll angle $\theta$ will be introduced into the imaging process. This fact leads to gradual disparity change in the horizontal direction (see Fig. \ref{fig.disp}), making the way of representing road disparity projections using (\ref{eq.linear_model}) somewhat problematic \cite{fan2018novel}. Furthermore, compared to the case that the roll angle is zero, the disparity distribution of each row becomes less compact and $E_\text{min}$ becomes much higher. Therefore, the roll angle has to be considered when minimizing (\ref{eq.E}). 

\begin{figure*}[!t]
	\centering
	\includegraphics[width=0.99999\textwidth]{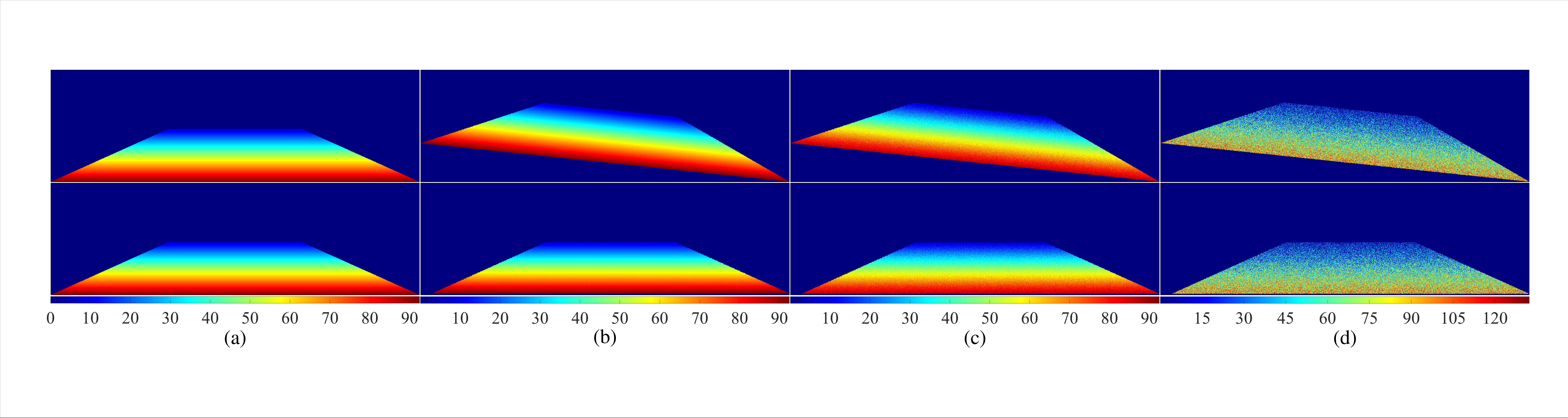}
	\caption{Experimental results of roll angle estimation: (a) $\theta=0^\circ$ and $\kappa=0$; (b) $\theta=10^\circ$ and $\kappa=0$; (c) $\theta=10^\circ$ and $\kappa=5$; (d) $\theta=10^\circ$ and $\kappa=40$;}
	\label{fig.roll_angle_exp}
		\vspace{-1em}
\end{figure*}
\begin{figure}[!t]
	\centering
	\includegraphics[width=0.38\textwidth]{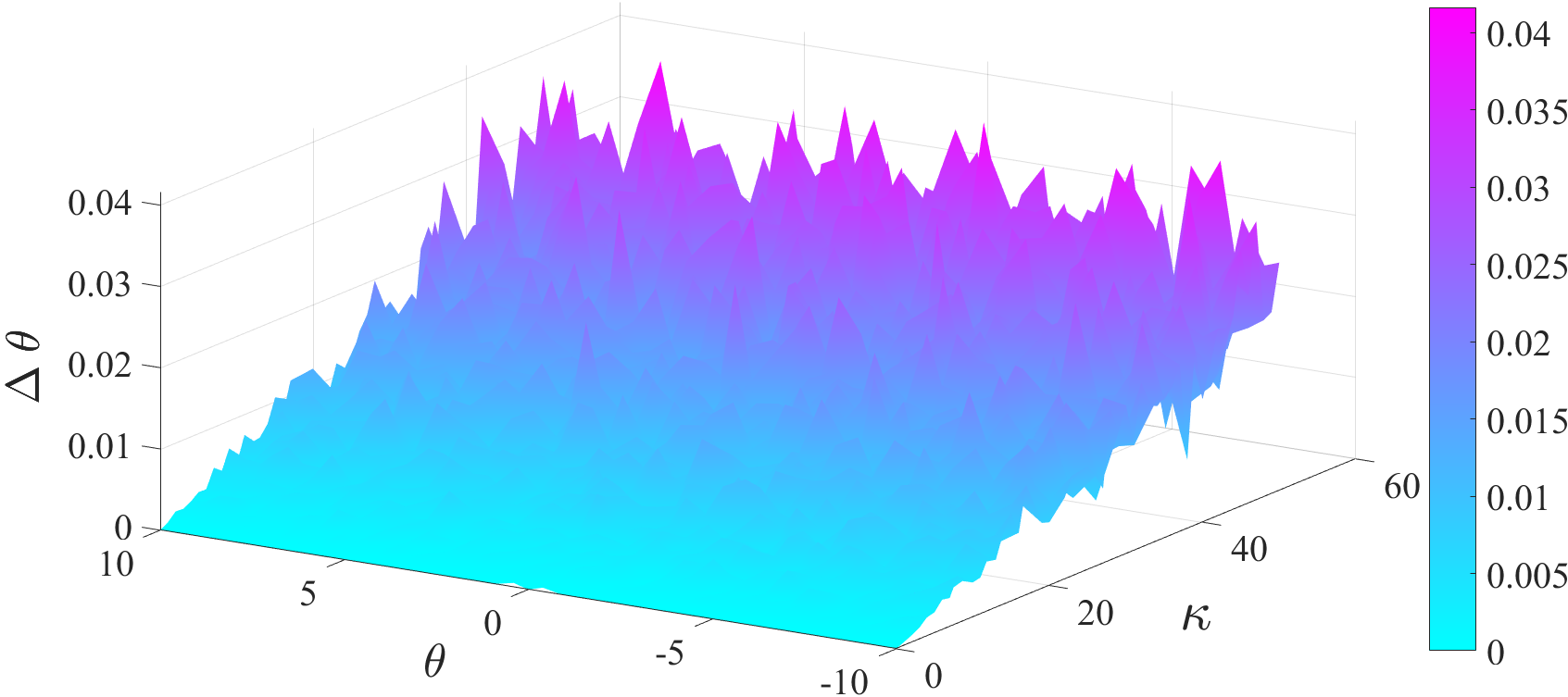}
	\caption{$\Delta\theta$ with respect to $\theta$ and $\kappa$. }
	\vspace{-1.5em}
	\label{fig.error_map}
\end{figure}
To rotate the disparity map at a given angle $\theta$ around the image center, each original point $\mathbf{p}=[u,v]^\top$ is transformed to a new point $\mathbf{q}=[s,t]^\top$ using:
\begin{equation}
\mathbf{q}(\theta, \mathbf{p})=\begin{bmatrix}
\cos\theta & \sin\theta\\
-\sin\theta & \cos\theta
\end{bmatrix}\mathbf{p}.
\label{eq.R}
\end{equation}
(\ref{eq.E}) can, therefore, have a more general expression:
\begin{equation}
E(\theta)=\norm{\mathbf{d}-\mathbf{T}(\theta)\mathbf{a}(\theta)}_2^2,
\label{eq.E2}
\end{equation}
where $\mathbf{T}(\theta)=[\mathbf{1}_{n}, \ \mathbf{t}(\theta)]$ and $\mathbf{t}=[t_1(\theta),t_2(\theta),\cdots,t_n(\theta)]^\top$. The closed form solution for (\ref{eq.E2}) is as follows:
\begin{equation}
\mathbf{a}(\theta)=\big(\mathbf{T}(\theta)^\top\mathbf{T}(\theta)\big)^{-1}\mathbf{T}(\theta)^{\top}\mathbf{d}.
\label{eq.alpha2}
\end{equation}
(\ref{eq.E_min}) can, therefore, be rewritten as follows:
\begin{equation}
E_\text{min}(\theta)
=\mathbf{d}^\top\mathbf{d}-\mathbf{d}^\top\mathbf{T}(\theta)\big(\mathbf{T}(\theta)^\top\mathbf{T}(\theta)\big)^{-1}\mathbf{T}(\theta)^\top\mathbf{d}.
\label{eq.E_min2}
\end{equation}
Minimizing (\ref{eq.E_min2}) is equivalent to maximizing: 
\begin{equation}
\begin{split}
g(\theta)=\mathbf{d}^\top\mathbf{T}(\theta)\big(\mathbf{T}(\theta)^\top\mathbf{T}(\theta)\big)^{-1}\mathbf{T}(\theta)^\top\mathbf{d}\ \ \text{s.t.}\ \theta\in(-\frac{\pi}{2},\frac{\pi}{2}].
\end{split}
\label{eq.g}
\end{equation}
\begin{figure*}[!t]
	\centering
	\includegraphics[width=0.99999\textwidth]{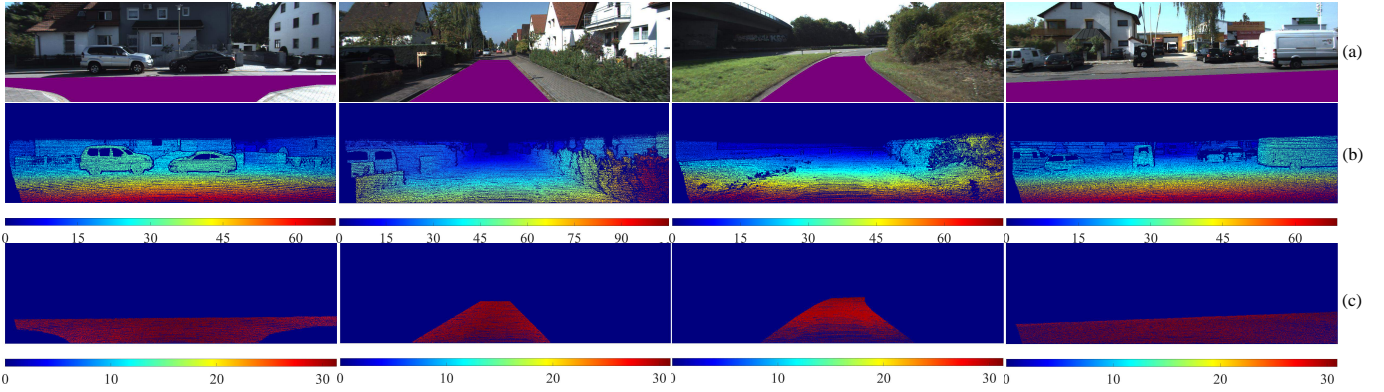}
	\caption{Experimental results of the KITTI stereo dataset: (a) left stereo images; (b) original disparity maps; (c) transformed disparity maps;}
	\vspace{-1em}
	\label{fig.kitti}
\end{figure*}
\begin{figure*}[!t]
	\centering
	\includegraphics[width=0.99999\textwidth]{./apollo.pdf}
	\caption{Experimental results of the ApolloScape stereo dataset: (a) left stereo images; (b) original disparity maps; (c) transformed disparity maps;}
	\vspace{-1em}
	\label{fig.apollo}
\end{figure*}
\begin{figure*}[!t]
	\centering
	\includegraphics[width=0.9999\textwidth]{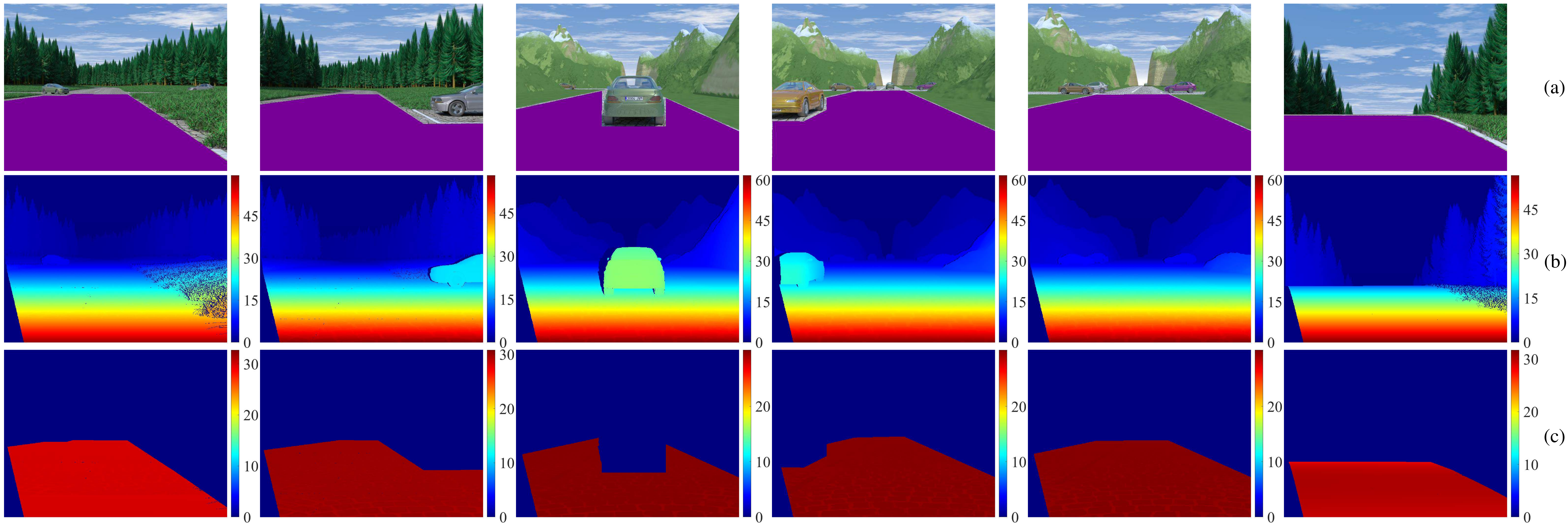}
	\caption{Experimental results of the EISATS stereo dataset: (a) left stereo images; (b) original disparity maps; (c) transformed disparity maps;}
	\label{fig.EISATS}
	\vspace{-1em}
\end{figure*}
According to (\ref{eq.R}), we can obtain:
\begin{equation}
\mathbf{F}=
\mathbf{T}(\theta)^\top \mathbf{T}(\theta)=\begin{bmatrix}
n & r_0(\theta)\\
r_0(\theta) & r_1(\theta)\\
\end{bmatrix},
\label{eq.F}
\end{equation}
where 
\begin{equation}
r_0(\theta)=\mathbf{v}^\top\mathbf{1}_{n}\cos\theta-\mathbf{u}^\top\mathbf{1}_{n}\sin\theta,
\label{eq.r0}
\end{equation}
\begin{equation}
r_1(\theta)= \frac{\mathbf{v}^\top\mathbf{v}
	+\mathbf{u}^\top\mathbf{u}}{2}
+
\frac{\mathbf{v}^\top\mathbf{v}
	-\mathbf{u}^\top\mathbf{u}}{2}\cos 2\theta
-\mathbf{u}^\top\mathbf{v} \sin 2\theta,
\label{eq.r1}
\end{equation}
$\mathbf{u}=[u_1,\ u_2,\ \cdots,\ u_n]^\top$ and $\mathbf{v}=[v_1,\ v_2,\ \cdots,\ v_n]^\top$ are two column vectors storing the horizontal and vertical coordinates, respectively. (\ref{eq.F}), (\ref{eq.r0}) and (\ref{eq.r1}) result in the following expression:
\begin{equation}
\mathbf{F}^{-1}=\frac{1}{nr_1(\theta)-{r_0(\theta)}^2}\begin{bmatrix}
r_1(\theta) & -r_0(\theta)\\
-r_0(\theta) & n
\end{bmatrix},
\label{eq.F_1}
\end{equation}
Plugging (\ref{eq.r0})-(\ref{eq.F_1}) into (\ref{eq.g}) results in the following expression:
\begin{equation}
g(\theta)=\frac{w_3+w_4 \cos 2\theta+w_5 \sin 2\theta}{w_0+w_1 \cos 2\theta+w_2 \sin 2\theta}\ \ \text{s.t.}\ \theta\in(-\frac{\pi}{2},\frac{\pi}{2}],
\label{eq.g1}
\end{equation}
where 
\begin{equation}
w_0=\frac{1}{2}\big[
n(\mathbf{v}^\top\mathbf{v}+\mathbf{u}^\top\mathbf{u})-
(\mathbf{v}^\top\mathbf{1}_{n})^2-(\mathbf{u}^\top\mathbf{1}_{n})^2
\big],
\end{equation}
\begin{equation}
w_1=\frac{1}{2}\big[
n(\mathbf{v}^\top\mathbf{v}-\mathbf{u}^\top\mathbf{u})-
(\mathbf{v}^\top\mathbf{1}_{n})^2+(\mathbf{u}^\top\mathbf{1}_{n})^2
\big],
\end{equation}
\begin{equation}
w_2=\mathbf{v}^\top\mathbf{1}_{n}\mathbf{u}^\top\mathbf{1}_{n}-n\mathbf{v}^\top\mathbf{u},
\end{equation}
\begin{equation}
\begin{split}
w_3=\frac{1}{2}\big[(\mathbf{d}^\top&\mathbf{1}_{n})^2(\mathbf{v}^\top\mathbf{v}+\mathbf{u}^\top\mathbf{u})+n\big((\mathbf{d}^\top\mathbf{v})^2+(\mathbf{d}^\top\mathbf{u})^2\big)\big]\\
&-\mathbf{d}^\top\mathbf{1}_{n}\big(\mathbf{v}^\top\mathbf{1}_{n}\mathbf{d}^\top\mathbf{v}+\mathbf{u}^\top\mathbf{1}_{n}\mathbf{d}^\top\mathbf{u}\big),
\end{split}
\end{equation}
\begin{equation}
\begin{split}
w_4=\frac{1}{2}\big[  
(\mathbf{d}^\top &\mathbf{1}_{n})^2 (\mathbf{v}^\top\mathbf{v}-\mathbf{u}^\top\mathbf{u})+
n\big(
(\mathbf{d}^\top\mathbf{v})^2-(\mathbf{d}^\top\mathbf{u})^2
\big)          
\big]\\
&-\mathbf{d}^\top\mathbf{1}_{n}
\big(\mathbf{v}^\top\mathbf{1}_{n}\mathbf{d}^\top\mathbf{v}-\mathbf{u}^\top\mathbf{1}_{n}\mathbf{d}^\top\mathbf{u}
\big), 
\end{split}
\end{equation}
\begin{equation}
\begin{split}
w_5=\mathbf{d}^\top\mathbf{1}_{n}\big(\mathbf{v}^\top\mathbf{1}_{n}\mathbf{d}^\top\mathbf{u}+\mathbf{u}^\top\mathbf{1}_{n}\mathbf{d}^\top\mathbf{v}\big)-(\mathbf{d}^\top\mathbf{1}_{n})^2\mathbf{v}^\top\mathbf{u}-n\mathbf{d}^\top\mathbf{v}\mathbf{d}^\top\mathbf{u}.
\end{split}
\end{equation}
The angle $\theta$ which maximizes $g(\theta)$ can be obtained by differentiating $g(\theta)$ with respect to $\theta$:
\begin{equation}
\begin{split}
\frac{\delta g(\theta)}{\delta \theta}=\frac{-2}{(w_0+w_1 \cos 2\theta+w_2 \sin 2\theta)^2}\Big((w_4w_2-w_5w_1)
\\+(w_3w_2-w_5w_0)\cos 2\theta
+(w_4w_0-w_3w_1)\sin 2\theta
\Big).
\label{eq.dif_g}
\end{split}
\end{equation}
If the denominator of  (\ref{eq.dif_g}) does not equal zero, we can get two angles  $\theta_1$ and $\theta_2$ at which $g(\theta)$ achieves the extrema: 
\begin{equation}
\begin{split}
\theta_1=\arctan\Big(\frac{w_4w_0-w_3w_1+\sqrt{\Delta}}{w_3w_2+w_5w_1-w_5w_0-w_4w_2} \Big),
\end{split}
\label{eq.theta1}
\end{equation}
\begin{equation}
\begin{split}
\theta_2=\arctan\Big(\frac{w_4w_0-w_3w_1-\sqrt{\Delta}}{w_3w_2+w_5w_1-w_5w_0-w_4w_2} \Big),
\end{split}
\label{eq.theta2}
\end{equation}
where:
\begin{equation}
\begin{split}
\Delta=(w_4w_0-w_3w_1)^2+(w_3w_2-w_5w_0)^2-(w_4w_2-w_5w_1)^2.
\end{split}
\label{eq.delta}
\end{equation}
The desirable roll angle $\theta$ can, therefore, be determined by finding the highest value between $g(\theta_1)$ and $g(\theta_2)$. $\mathbf{a}$ can then be obtained by substituting $\theta$ into (\ref{eq.alpha2}). Each road disparity can now be represented using:
\begin{equation}
f(\mathbf{p},\theta)=a_0+a_1(-u\sin\theta  +v\cos\theta ).
\label{eq.f}
\end{equation}
Damaged and undamaged road areas can now be better distinguished by transforming the original disparity map $D$ (see Fig. \ref{fig.disp}) to a new disparity map $\tilde{D}$ (see Fig. \ref{fig.trans_disp}) using:
\begin{equation}
\tilde{D}(\mathbf{p})=D(\mathbf{p})-f(\mathbf{p},\theta)+\delta,
\label{eq.disp_trans} 
\end{equation}
where $\delta$ can be any constant enabling the transformed disparity values to be non-negative. The transformed disparity map is shown in Fig. \ref{fig.trans_disp}. Finally, the damaged road areas can be extracted by applying Otsu's thresholding method on the transformed disparity map. The corresponding result is shown in Fig. \ref{fig.seg}. 

\begin{figure*}[!t]
	\centering
	\includegraphics[width=0.90\textwidth]{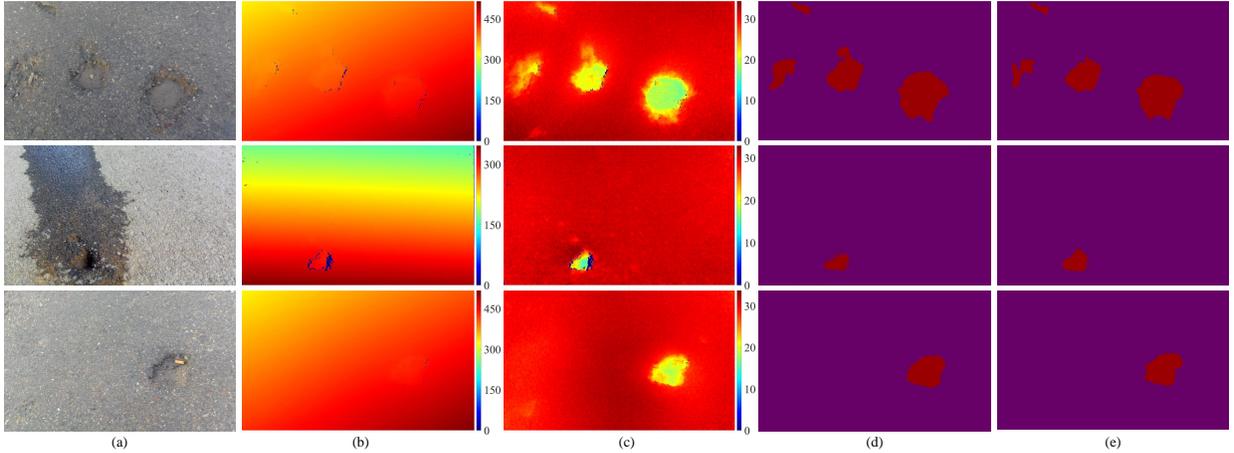}
	\caption{Experimental results of road damage detection: (a) left stereo images; (b) original disparity maps; (c) transformed disparity maps; (d) detection results; (e) ground truth; the regions in purple and red are undamaged and damaged road areas, respectively.}
	\vspace{-1em}
	\label{fig.experimental_results}
\end{figure*}
\section{Experimental Results}
\label{sec.experimental_results}
The proposed road damage detection algorithm is programmed in both C++ and Matlab C on an Intel Core i7-8700K CPU (3.7 GHz) using a single thread. The remainder of this section discusses the performance of roll angle estimation, disparity transformation, and road damage detection, respectively.

\subsection{Roll Angle Estimation Evaluation}
\label{sec.roll_angle_estimation_evaluation}
To quantify the accuracy the proposed roll angle estimation algorithm, we created a synthetic dataset (including 51 dense disparity maps with respect to different roll angles $\theta\in[-10^\circ, +10^\circ]$). The KITTI  stereo rig  configurations\footnote{http://www.cvlibs.net/datasets/kitti/setup.php} are utilized to create these synthetic disparity maps.  To further evaluate the robustness of the proposed roll angle estimation algorithm, we added Gaussian white noise $\kappa\omega$ to the synthetic disparity maps, where $\omega\in[-1, +1]$ is a random decimal value and $\kappa$ is a scale parameter set to control the intensity of the noise. Some examples of the experimental results are illustrated in Fig. \ref{fig.roll_angle_exp}, where the first row illustrates the original disparity maps and the second row shows the disparity maps rotated around the estimated roll angles.  To quantify the accuracy of our proposed roll angle estimation algorithm, we compute the absolute difference $\Delta \theta$ between the estimated and actual roll angles, i.e., $\tilde{\theta}$ and $\theta$. $\Delta \theta$ with respect to different $\theta$ and $\kappa$ is shown in Fig. \ref{fig.error_map}, where we can observe that the accuracy of our proposed roll angle estimation algorithm decreases with the increase of $\kappa$, but the highest $\Delta \theta$ is only about $0.04^\circ$ ($\kappa=50$). Therefore, our proposed roll angle estimation algorithm is highly accurate and very robust to noise.

\subsection{Disparity Transformation Evaluation}
\label{sec.disp_trans_evaluation}
As discussed in Section \ref{sec.algorithm_description}, the road damage becomes highly distinguishable after unsupervised disparity transformation. The transformed disparities in the undamaged road areas tend to have similar values, while they differ greatly from those in the damaged road areas, as shown in Fig. \ref{fig.trans_disp}.  

In our experiments, we utilized the KITTI stereo \cite{Geiger2012, Menze2015a}, the ApolloScape\footnote{http://apolloscape.auto/stereo.html} stereo, and the EISATS stereo \cite{Vaudrey2008, Wedel2008} datasets to evaluate the performance of our proposed disparity transformation algorithm. The former two datasets are used for the evaluation of sparse and dense real-world disparity map transformation, respectively. The EISATS stereo dataset is utilized to evaluate the performance of our proposed algorithm on 
synthetic disparity maps. The corresponding experimental results are illustrated in Fig. \ref{fig.kitti}, \ref{fig.apollo} and \ref{fig.EISATS}, respectively, where the areas in purple are our manually labeled road regions. 

To quantify the  disparity transformation accuracy, we introduced a measure named transformed disparity standard deviation $\sigma$:
\begin{equation}
\sigma=\sqrt{\frac{1}{m}\norm{\tilde{\mathbf{d}}-\frac{\tilde{\mathbf{d}}^\top\mathbf{1}_{m}}{m}}_2^2},
\end{equation}
where $\tilde{\mathbf{d}}=[\tilde{D}(\mathbf{p}_1), \tilde{D}(\mathbf{p}_2), \cdots, \tilde{D}(\mathbf{p}_m)]^\top$ stores the transformed disparity values. We compare our proposed method with GSS-DP \cite{fan2018novel} and GD \cite{fan2019real}. The comparisons of $\sigma$ and runtime are illustrated in Table \ref{table.mu_sigma}. It can be clearly seen that our proposed algorithm achieves the minimum $\sigma$ on all the stereo datasets. Furthermore, as our proposed algorithm can directly obtain the numerical solution for (\ref{eq.E2}), it performs much faster than both \cite{fan2018novel} and \cite{fan2019real}. 
\begin{table}[!h]
	\begin{center}
		\vspace{0in}
		\footnotesize
		\caption{Comparisons of $\sigma$ and runtime. }
		\label{table.mu_sigma}
		\begin{tabular}{|c|c|c|c|c|}
			\hline
			Dataset & Method  & $\sigma$ & runtime (ms) \\
			\hline
			\multirow{3}{*}{KITTI} & GSS-DP \cite{fan2018novel}   & 0.4305 & 32.3182 \\
			\cline{3-4}
			& GD \cite{fan2019real} & 0.4299 & 7.5107 \\
			\cline{3-4}
			& Proposed   & \textbf{0.4289} & \textbf{1.1279} \\
			\hline
			\multirow{3}{*}{ApolloScape} & GSS-DP \cite{fan2018novel}   & 0.2614 & 523.5302
			\\
			\cline{3-4}
			& GD \cite{fan2019real}  & 0.2613 & 91.8474
			\\
			\cline{3-4}
			& Proposed   & \textbf{0.2610}  & \textbf{37.2680} \\
			\hline

			\multirow{3}{*}{EISATS} & GSS-DP \cite{fan2018novel}  & 0.2757 & 45.6317\\
			\cline{3-4}
			& GD \cite{fan2019real}  & 0.2752 &8.9177 \\
			\cline{3-4}
			& Proposed   & \textbf{0.2703} & \textbf{2.3574} \\
			\hline
		\end{tabular}
	\end{center}
\end{table}
\vspace{-1.5em}

\subsection{Road Damage Detection Evaluation}
\label{sec.damage_detection_evaluation}
\begin{figure}[!t]
	\centering
	\includegraphics[width=0.30\textwidth]{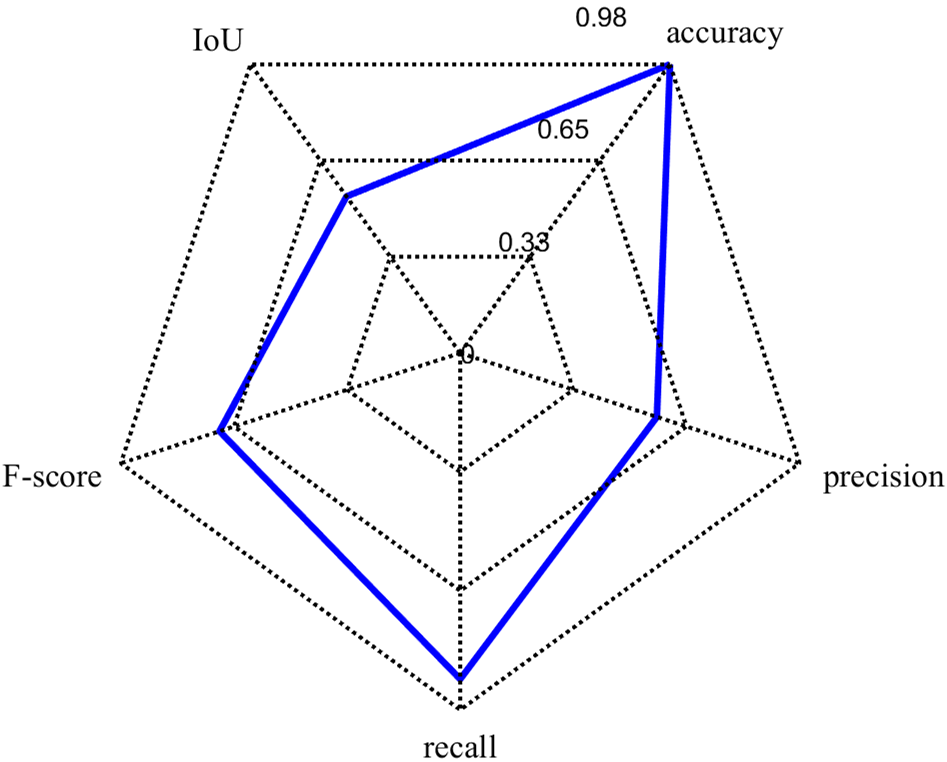}
	\caption{The pixel-level accuracy, precision, recall, F-score,  and IoU  achieved using the proposed algorithm. }
	\label{fig.radar}
		\vspace{-1.5em}
\end{figure}

In this subsection, we utilize our recently published pothole detection dataset\footnote{\url{ruirangerfan.com}}  \cite{fan2019pothole} to evaluate the performance of road damage detection. Some examples of the detected damaged road areas are shown on the fourth column in Fig. \ref{fig.experimental_results}.
To quantify the accuracy of our proposed road damage detection algorithm, we compute the pixel-level precision, recall, F-score, IoU, and accuracy, as shown in Fig. \ref{fig.radar}. It can be seen that our proposed road damage detection algorithm performs  accurately.  The pixel-level accuracy of the detected road damage areas is approximately $97.56\%$.
\vspace{-1em}
\section{Conclusion}
\label{sec.conclusion}
This paper presented a novel road damage detection algorithm based on unsupervised disparity map segmentation. This was achieved by minimizing an energy function with respect to the stereo rig roll angle and the road disparity projection model. Instead of minimizing this energy function using non-linear optimization methods, such as GSS-DP and GD, we directly found its numerical solution, which enables our proposed algorithm to perform more accurately and efficiently than GSS-DP and GD. A dense disparity map can, therefore, be transformed to better distinguish between damaged and undamaged road areas. By applying Otsu's thresholding method on the transformed disparity map, the road damage can then be effectively detected. The proposed algorithm does not require any parameters when transforming and segmenting road disparity maps. The experimental results also demonstrated that our algorithm can perform in real time. The pixel-level accuracy of the detected road damage areas is approximately $97.56\%$.


\bibliographystyle{IEEEtran}

\end{document}